\documentclass[final]{cvpr}

\usepackage{times}
\usepackage{epsfig}
\usepackage{graphicx}
\usepackage{amsmath}
\usepackage{amssymb}
\usepackage{booktabs}
\usepackage{multirow}

\graphicspath{ {./fig/} }

\usepackage[pagebackref=true,breaklinks=true,colorlinks,bookmarks=false]{hyperref}

\def\cvprPaperID{2063} 
\def\confYear{CVPR 2021}

\newcommand*{\affaddr}[1]{#1} 
\newcommand*{\affmark}[1][*]{\textsuperscript{#1}}

\begin{document}

\title{Context-Aware Layout to Image Generation with Enhanced Object Appearance}
\makeatletter
\newcommand{\printfnsymbol}[1]{%
  \textsuperscript{\@fnsymbol{#1}}%
}
\makeatother


\author{%
Sen He\affmark[1,2\thanks{Equal contribution}], Wentong Liao\affmark[3\printfnsymbol{1}], Michael Ying Yang\affmark[4], 
Yongxin Yang\affmark[1,2], Yi-Zhe Song\affmark[1,2]
\\ Bodo Rosenhahn\affmark[3], Tao Xiang\affmark[1,2]
\\
\normalsize
\affaddr{\affmark[1]CVSSP, University of Surrey},
\affaddr{\affmark[2]iFlyTek-Surrey Joint Research Centre on Artificial Intelligence}\\
\normalsize
\affaddr{\affmark[3]TNT, Leibniz University Hannover}, 
\affaddr{\affmark[4]SUG, University of Twente}\\
\normalsize
\vspace{-1em}
}
\date{}

\maketitle

\begin{abstract}
A layout to image (L2I) generation model aims to generate a complicated image containing multiple objects (things) against natural background (stuff), conditioned on a given layout. Built upon the recent advances in generative adversarial networks (GANs), existing L2I models have made great progress. However, a close inspection of their generated images reveals two major limitations: (1) the object-to-object as well as object-to-stuff relations are often broken and (2) each object's appearance is typically distorted lacking the key defining characteristics associated with the object class. We argue that these are caused by the lack of context-aware object and stuff feature encoding in their generators, and location-sensitive appearance representation in their discriminators.  To address these limitations, two new modules are proposed in this work. First, a context-aware feature transformation module is introduced in the generator to ensure that the generated feature encoding of either object or stuff is aware of other co-existing objects/stuff in the scene. Second, instead of feeding location-insensitive image features to the discriminator, we use the Gram matrix computed from the feature maps of the generated object images to preserve location-sensitive information, resulting in much enhanced object appearance. 
Extensive experiments show that the proposed method achieves state-of-the-art performance on the COCO-Thing-Stuff and Visual Genome benchmarks. Code available at: \url{https://github.com/wtliao/layout2img}.

\end{abstract}

\begin{figure}[t]
\centering
\includegraphics[width=0.4\textwidth]{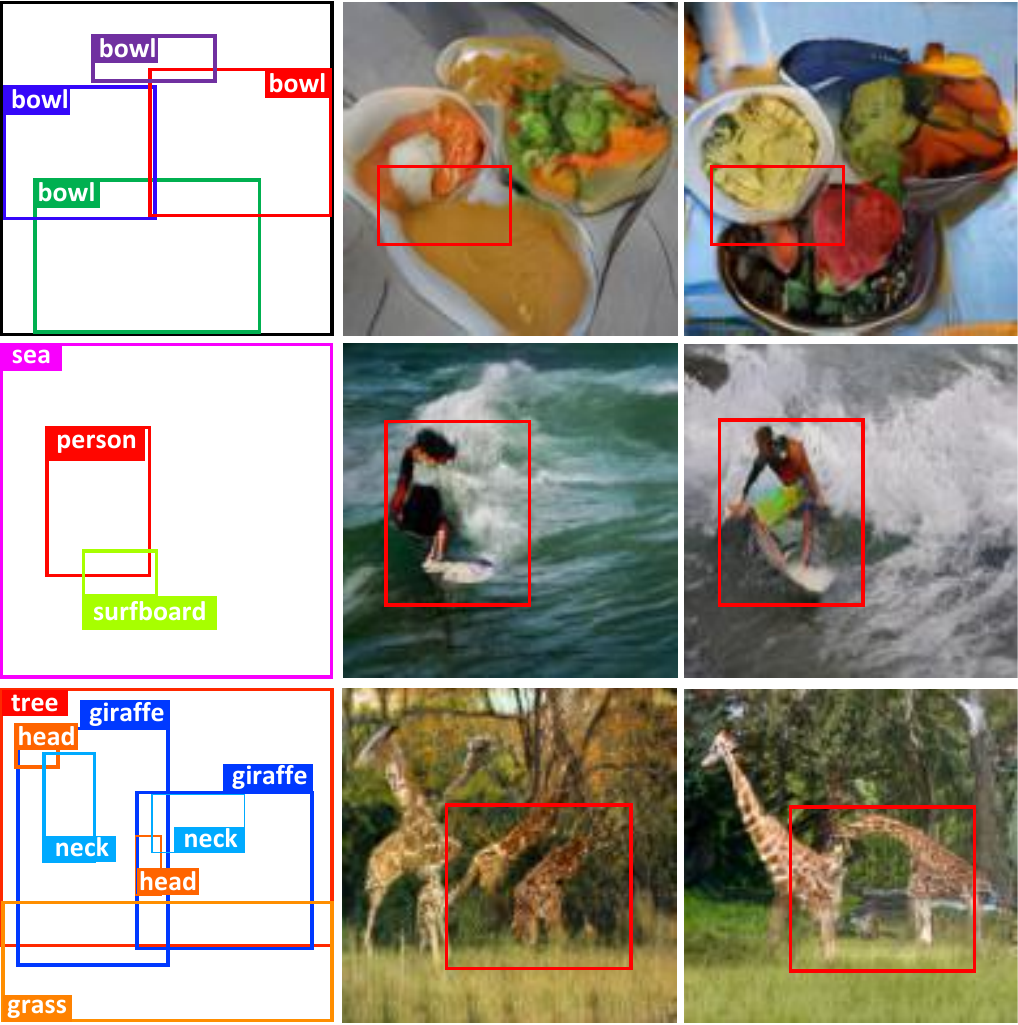}
~
\caption{\small Illustration of the limitations of existing L2I models and how our model overcome them. From left to right: ground truth layout, images generated  by the state-of-the-art LostGAN-v2 \cite{sun2020learning}, and by our model with the layout as input. In the middle and right column, regions with key differences in the generation quality between LostGAN-v2 and our model are highlighted in dashed boxes. See text for more details. 
}
\vspace{-10pt}
\label{fig:topic}
\end{figure}
\vspace{-0.6cm}
\section{Introduction}

Recent advances in generative adversarial networks (GANs) \cite{goodfellow2014generative} have made it possible to generate photo-realistic images for a single object, e.g., faces, cars, cats \cite{brock2018large,zhang2019self, karras2019style,karras2020analyzing}. However, generating complicated images containing  multiple objects (things) of different classes against natural backgrounds (stuff) still remains a challenge \cite{johnson2018image,ashual2019specifying,reed2016generative,park2019semantic}. This is due to the large appearance variations for objects of different classes, as well as the complicated relations between both object-to-object and object-to-stuff. A generated object needs to be not only realistic on its own, but in harmony with surrounding objects and stuff. 

Without any conditional input, the \textit{mode collapse} \cite{salimans2016improved,che2016mode} problem is likely to be acute for GANs trained to generate such complicated natural scenes. Consequently, various inputs have been introduced to provide some constraints on the image generation process. These include textual description of image content \cite{reed2016generative}, scene graph representing objects and their relationship \cite{johnson2018image}, and semantic map providing pixel-level annotation \cite{park2019semantic}. This work focuses on the conditional image generation task using the layout \cite{zhao2019image,sun2019image,sylvain2020object} that defines a set of bounding boxes with specified size, location and categories  (see Fig.~\ref{fig:topic}). Layout is a user-friendly input format on its own and can also be used as an intermediate input step of other tasks, e.g., scene graph and text to image generation \cite{ashual2019specifying,hong2018inferring}.

Since the seminal work \cite{zhao2019image} in 2019, the very recent layout to image (L2I) generation models \cite{zhao2020layout2image,sylvain2020object,sun2020learning}  have made great progresses, thanks largely to the advances made in GANs \cite{park2019semantic,karras2019style} as they are the key building blocks. From a distance, the generated images appear to be realistic and adhere to the input layout (see Fig.~\ref{fig:topic} and more in Fig.~\ref{fig:qualitative}). However, a closer inspection reveals two major limitations. First, the relations between objects and object-to-stuff are often broken. This is evident from the food example in Fig.~\ref{fig:topic} (Top-Middle) -- the input layout clearly indicates that the four bowls are overlapping with each other. Using the state-of-the-art LostGAN-v2 \cite{sun2020learning}, the occluded regions between objects are poorly generated.  Second, each generated object's appearance is typically distorted lacking class-defining characteristics. For instance, the surfing example in  Fig.~\ref{fig:topic} (Middle) and the giraffe example in  Fig.~\ref{fig:topic} (Bottom) show that the object appearance  has as if been touched by Picasso -- one can still recognize the surfing person or giraffe, but key body parts are clearly misplaced. 


We believe these limitations are caused by two major design flaws in existing L2I models in both their GAN generators and discriminators. (1) \textit{Lack of context-aware modeling in the generator}:  Existing models generate the feature for the object/stuff in each layout bounding box first, and then feed the generated feature into a generator for image generation.  However, the feature generation process for each object/stuff is completely independent of each other, therefore offering no chance for capturing the inter-object and object-to-stuff relations.  (2) \textit{Lack of location-sensitive appearance representation in the discriminator}: As in any GAN model, existing L2I models deploy a discriminator that is trained to distinguish the generated whole image and individual object/stuff images from the real ones. Such a discriminator is essentially a CNN binary classifier whereby globally pooled features extracted from the CNN are fed to a real-fake classifier. The discriminator thus cares only about the presence/absence and strength of each semantic feature, rather than where they appear in the generated images. This lack of location-sensitive appearance representation thus contributes to the out-of-place object part problem in  Fig.~\ref{fig:topic} (Middle). 

In this paper, we provide solutions to overcome both limitations. First, to address the lack of context-aware modeling problem,  we propose to introduce a context-aware feature transformation module in the generator of a L2I model. This module updates the generated feature for each object and stuff after each has examined its relations with all other objects/stuff co-existing in the image through self-attention. Second, instead of feeding location-insensitive globally pooled object image features to the discriminator, we use the Gram matrix computed from the feature maps of the generated object images. The feature map Gram matrix captures the inter-feature correlations over the vectorized feature map, and is therefore locations sensitive.  Adding it to the input of the real-fake classifier in the discriminator, the generated images preserve both shape and texture characteristics of each object class, resulting in much enhanced object appearance (see Fig.~\ref{fig:topic} (Right)).

%

\textbf{The contributions} of this work are as follows: (1) For the first time, we identify two major limitations of existing L2I models for generating complicated multi-object images. (2) Two novel components, namely a context-aware feature transformation module and a location-sensitive object appearance representation are introduced to address these two limitations. (3) The proposed modules can be easily integrated into any existing L2I generation models and improve them significantly. (4) Extensive experiments on both the COCO-Thing-Stuff \cite{lin2014microsoft,caesar2018coco} and Visual Genome \cite{krishna2017visual} datasets show that state-of-the-art performance is achieved using our model. The code and trained models will be released soon. 

\section{Related Work}

\paragraph{Generative Adversarial Networks}
Generative adversarial networks (GANs) \cite{goodfellow2014generative}, which play a min-max game between a generator and a discriminator, is the mainstream approach used in recent image generation works. However, the training of a GAN is often unstable and known to be prone to the \textit{mode collapse} problem. To address this, techniques like Wasserstein GAN \cite{arjovsky2017wasserstein} and Unrolled GAN \cite{metz2016unrolled} were developed. Meanwhile, noise injection and weight penalizing \cite{arjovsky2017towards,roth2017stabilizing} were used in the discriminator to alleviate the non-convergence problem for further stabilization of the training. To generate high fidelity and resolution images, architectures like Progressive GAN  \cite{karras2017progressive} and BigGAN \cite{brock2018large} were also proposed.
\vspace{-0.4cm}

\paragraph{Conditional Image Generation}
Conditional image generation, which generates an image based on a given condition (e.g., class label, sentence description, image, semantic mask, sketch, and scene graph) has been studied intensively \cite{mirza2014conditional,reed2016generative,isola2017image, zhu2017unpaired,park2019semantic,chen2020deepfacedrawing,ashual2019specifying,gao2020sketchycoco} due to its potential in generating complicated natural images. In general, there are two popular architectures for the conditional image generation. The first one is the encoder-decoder architecture used in \textit{Pix2pix} \cite{isola2017image} and \textit{CycleGAN} \cite{zhu2017unpaired}, where the encoder directly takes the conditional input and embeds it to a latent space.  The decoder then transfers the embedded representation into the target image. 
The second popular architecture is the decoder-only architecture used in \textit{StyleGAN} \cite{karras2019style} and \textit{GauGAN} \cite{park2019semantic}, where a decoder starts with a random input, and then progressively transforms it to produce the desired output. In this architecture, the conditional input is used to generate part of the parameters in the decoder, e.g., the affine transformation parameters in the normalization layers \cite{karras2019style,park2019semantic,richardson2020encoding} or the weight parameters in convolutional kernels \cite{liu2019learning}.
\vspace{-0.4cm}

\paragraph{Layout to Image Generation}
Though the previous work \cite{hong2018inferring} has already touched the concept of layout to image generation  (L2I), it is just used as an intermediate step for a different generation task. The first stand-alone solution appeared in  \cite{zhao2019image}. Compared to other conditional inputs such as text and scene graph, layout is a more flexible and richer format. Therefore, more studies followed up by introducing more powerful generator architectures \cite{sun2019image,sun2020learning}, or new settings \cite{li2020bachgan,ma2020attribute}. Sun~\etal \cite{sun2019image} proposed a new architecture inspired by \textit{StyleGAN} \cite{karras2019style}, which allows their model to generate higher resolution images with better quality. Li~\etal \cite{li2020bachgan} introduced a new setting for high resolution street scene generation. Their model retrieves a background from a database based on the given foreground layout. Recently, Ma~\etal \cite{ma2020attribute} introduced attribute guided layout generation, which is more controllable on the generated objects. 
As mentioned earlier, all these existing models have two limitations, namely lack of context-aware modeling in their generators, and lack of location-sensitive appearance representation in their discriminators. Both limitations are overcome in this work, resulting in much improved L2I generation performance (see Sec.~\ref{sec:exp}). 
\vspace{-0.4cm}

\paragraph{Context Modeling}
Context plays an important role in many discriminative scene analysis tasks~\cite{torralba2003context, hu2018relation,chen2018context, yao2018exploring, xu2019spatial,herdade2019image, wang2020robust}. The main idea in context-based analysis is to tie each object instance in the scene with the global context, such that their relationship or interaction can be better understood. However, context has drawn little attention in image generation. One exception is SAGAN \cite{zhang2019self} which applied self-attention to refine the feature map in the generator for single object image generation. In this work, we introduce context modeling for layout to image generation, a more complicated image generation task with a focus on inter-objects and object-to-stuff relation modeling. 
\vspace{-0.4cm}

\paragraph{Appearance Representation in CNNs}
Works on CNN visualization clear show that feature channels, especially those at the top layers of a CNN capture semantically meaningful concepts such as body parts;  and the activations of these feature channel at different locations indicate where these concepts are \cite{zhou2014object}. However, when it comes to object recognition \cite{ILSVRC15} or real-fake discriminator in GAN \cite{goodfellow2014generative}, these feature maps are globally  pooled before being fed into a binary classification layer. Location-sensitive information is thus largely lost, and the focus is on the presence/absence of the semantic concepts rather than where.  We therefore propose to use the Gram matrix computed on the feature maps to complement the semantics-only appearance representation used in existing discriminators in order to induce location-sensitivity in object image generation.  Such a Gram matrix based appearance representation has been used in    
style transfer \cite{gatys2015neural} for style/texture representation, which seems to suggest that it only captures feature distribution but contains no spatial information. However, as pointed out in \cite{Li2017DemystifyingNS}, this is because the use of entry-wise mean-square distance in \cite{gatys2015neural} removes the location sensitivity in the feature map Gram matrix. In our model, we pass the raw matrix instead of mean-square distance to the discriminator classifier, therefore preserving the location sensitivity.  


\section{Preliminaries}

\begin{figure*}[t]
\centering
\includegraphics[width=0.9\textwidth]{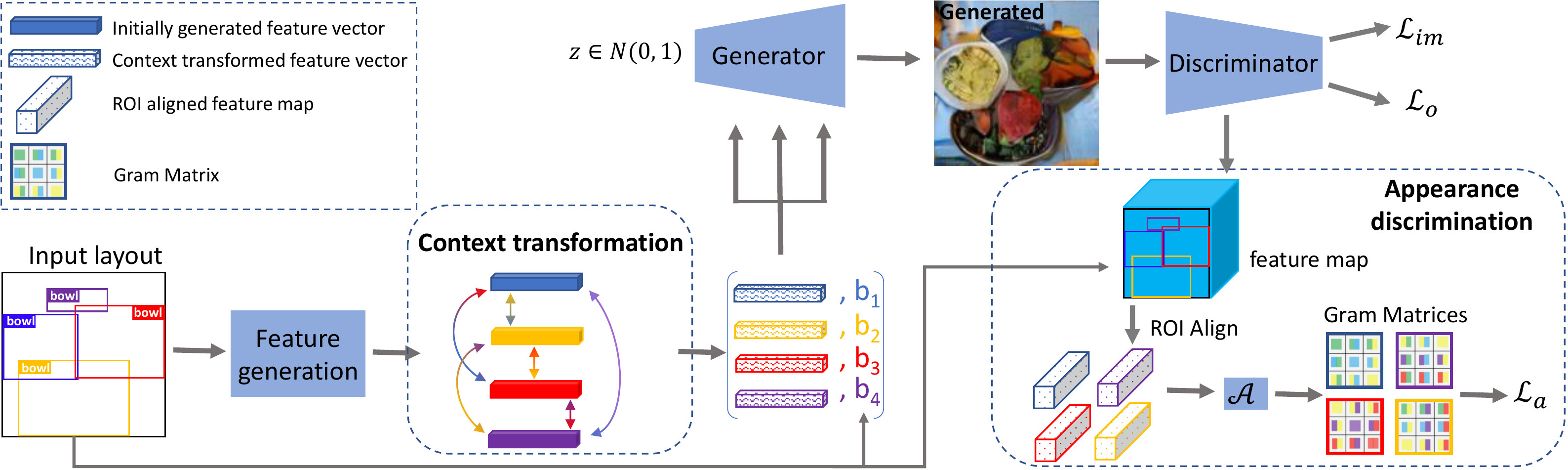}
~
\caption{\small A schematic of our method with a decoder-only generator as in \cite{sun2019image,zhao2020layout2image}. The feature generation module generates the raw representation for each bounding box based on their class label, the context-aware feature transformation module integrates the global context into the representation of each bounding box. Then the transformed bounding boxes' representation and the box coordinates ($b_i$) are fed into the generator for image generation. Finally the generated image is compared with real images by a discriminator with three losses, namely image-level and object-level semantic loss ($\mathcal{L}_{im}$ and $\mathcal{L}_{o}$) and object-level Gram matrix loss ($\mathcal{L}_{a}$).}
\label{fig:pipeline}
\vspace{-10pt}
\end{figure*}

\subsection{Problem Definition}
Let $L=\{ (y_{i},b_{i})_{i=1}^{N}\}$ be a layout with $N$ bounding boxes, where $y_{i} \in \mathcal{C}$ is the class of the bounding box and $b_{i} = [x_{i},y_{i},w_{i},h_{i}]$ is the position and size of the bounding box in the image lattice ($ H \times W$). The goal of the layout to image (L2I) generation task is to build a model $\mathcal{G}$, which can generate a realistic photo $I_{g}\in \mathbb{R}^{3\times H \times W}$, given the coarse information in the layout $L$.

\subsection{Prior Models}
Before introducing our proposed method in Sec.~\ref{sec:method}, we first briefly describe prior L2I models. In all previous models, the first step is always to generate a feature representation for each bounding box based on their classes:
\begin{equation}
    \mathbf{p}_{i} = \phi_{0} ([\mathbf{e}_{i},\mathbf{n}_{i}]),
\end{equation}
where $\mathbf{p_{i}} \in \mathbb{R}^{d_{l}+d_{n}}$ is the feature representation of the $i^{th}$ bounding box in the layout, $\phi_{0}$ is a linear transformation layer, $\mathbf{e_{i}} \in \mathbb{R}^{d_{l}}$ is the label embedding of $y_{i}$, and $\mathbf{n_{i}} \in \mathbb{R}^{d_{n}}$ is a random noise sampled from a zero-mean unit-variance multivariate Gaussian distribution. The generated feature vector set $\{\mathbf{p}_{i}\}_{i=1}^{N}$ is then fed into a generator $\mathcal{G}$ for image generation. Depending on how the generator uses the feature vector set to generate the image, the existing models can be grouped into the following two categories.
\vspace{-0.4cm}

\paragraph{L2I Models with Encoder-Decoder Generators}
These models deploy an  encoder-decoder generator \cite{zhao2019image,ma2020attribute} which takes the feature vector set as input, and then transfers the feature vector set into a sequence of feature maps. Each feature map is generated by filling the corresponding feature vector into the region in the image lattice based on their bounding box. The generated feature maps are then fed into an encoder, which embeds each feature map into a latent space separately. Those embedded feature maps are merged into a single one through a convolutional LSTM network \cite{xingjian2015convolutional}. Finally, a decoder transforms the combined feature into the target image. Mathematically, the encoder-decoder based method can be formulated as:
\begin{equation}
    I_{g} = \textbf{D}(\text{cLSTM}( \textbf{E}(\{\mathcal{F}(\textbf{p}_{i},b_{i})\}_{i=1}^{N}))),
\end{equation}
where $\mathcal{F(\cdot,\cdot)}$ is a filling operation, $\textbf{E}$ is the encoder, $\text{cLSTM}$ is the convolutional LSTM network, and $\textbf{D}$ is the decoder.
\vspace{-0.4cm}

\paragraph{L2I Models with Decoder-Only Generators}
These models \cite{sun2019image,sun2020learning,sylvain2020object} use a decoder-only generator to first generate an auxiliary mask\footnote{The mask is not a strictly binary mask, as it is the output of a layer with sigmoid activation.} for each bounding box for a fine-grained shape or structure prediction:
\begin{equation}\label{eq:mask}
   \mathcal{M}_{i} = \mathcal{R_{S}}(\psi(\textbf{p}_{i}),b_{i}), 
\end{equation}
where $\psi$ is a small convolutional neural network,  $\psi(\textbf{p}_{i}) \in \mathbb{R}^{H\times W}$, and $\mathcal{R_{S}}(\cdot,\cdot)$ is a resize operator, which resizes each generated mask and fit it to the corresponding region in the image lattice via up/down sampling. Then the decoder receives a zero-mean unit-variance multivariate random noise $\textbf{n}_{0}\in\mathbb{R}^{C_{0}\times H_{0} \times W_{0}}$ as input, and decode it into the target image by modulating the affine transformation in the normalization layer:
\vspace{-3mm}
\begin{equation}\label{eq:c_mask}
    \hat f_{l}  = \text{BatchNorm}(f_{l},\varphi_{l}(\sum_{i=1}^{N}\textbf{p}_{i}\otimes \mathcal{M}_{l_{i}})),
\end{equation}
where $\hat f_{l}$ and $f_{l}$ are the feature maps before and after normalization at the $l^{th}$ layer in the decoder, $\varphi_{l}$ is a small convolutional block to generate the pixel-wise affine transformation parameters,  $\mathcal{M}_{l_{i}}$ is the resized version of $\mathcal{M}_{i}$ to match the corresponding feature map's scale, and $\otimes$ is the outer product, by which a vector $\textbf{p}_{i}$ and a matrix $\mathcal{M}_{l_{i}}$ produce a 3D tensor.

\section{The Proposed Method}
\label{sec:method}

The main architecture of our proposed method is illustrated in Fig.~\ref{fig:pipeline}. The proposed  context-aware feature transformation module and location-sensitive Gram matrix based object appearance representation are integrated into the generator and discriminator respectively of a decoder-only L2I generation architecture \cite{sun2019image,sun2020learning,sylvain2020object}. Similarly they can be easily integrated with those employing an encoder-decoder architecture \cite{zhao2019image,ma2020attribute}.

\subsection{Context-Aware Feature Generation}
Let us first look at the feature transformation module. It is clear that the prior models process each bounding box independently (either in the feature generation stage or the mask generation stage in the decoder-only methods), disrespecting the other objects and stuff in the scene. As a result, the generated objects do not appear in harmony with  other co-existing objects and stuff in the scene and often appear to be out of place (see Fig.~\ref{fig:topic} and Fig.~\ref{fig:qualitative}).   To overcome this limitation, we propose a context-aware transformation module, which integrates contextual information into the feature representation of each bounding box by allowing each feature to cross-examine all other features via self-attention \cite{NIPS2017_3f5ee243}. Concretely, the contextualized representation of each bounding box is computed as:
\vspace{-3mm}
\begin{equation}
    \mathbf{p}_{i}^{c} = \sum_{j=1}^{N}w_{i,j}\mathbf{p}_{j}\mathbf{W}_{v},
\end{equation}
\vspace{-3mm}
\begin{equation}
    w_{i,j} = \frac{\text{exp}(\alpha_{i,j})}{\sum_{k=1}^{N}\text{exp}(\alpha_{i,k})},
\end{equation}
\begin{equation}
    \alpha_{i,j} = (\mathbf{p}_{i}\mathbf{W}_{q})(\mathbf{p}_{j}\mathbf{W}_{k})^{T},
\end{equation}
where $\mathbf{W}_{q},\mathbf{W}_{k} \ \text{and} \ \mathbf{W}_{v} \in \mathbb{R}^{(d_{l}+d_n)\times(d_{l}+d_{n})}$ are linear transformation layers. With the transformation, the contextualized representation of each bounding box not only has its own information, but also the global context in the layout. It is thus able to avoid the poor occlusion region generation problem shown in Fig.~\ref{fig:topic} (Top-Middle). Note that this module can be used for feature map filling in the encoder-decoder based methods, as well as the  mask generation and the feature modulation steps in the decoder-only methods. The contextualized feature representation is then fed into the generator for image generation (see Fig.~\ref{fig:pipeline}). 

\subsection{Location-Sensitive Appearance Representation}
To address the issue of lacking location-sensitive appearance representation in the discriminators of existing L2I models, we introduce a feature map Gram matrix based appearance representation. In existing models' discriminators, the input image $I_{im}$ is first processed by a convolutional neural network $\psi_{D}$, and represented as $f_{im} \in \mathbb{R}^{C \times H_{D}\times W_{D}}$:
\begin{equation}
    f_{im} = \psi_{D}(I_{im}).
\end{equation}
Existing L2I models then apply two losses in the discriminator to train the whole model: an image-level loss $\mathcal{L}_{im}$ according to the globally pooled feature of $f_{im}$, and an object-level conditional loss $\mathcal{L}_{o}$ based on the  ROI pooled \cite{ren2015faster} feature of each object in the image, concatenated with its corresponding class information. These losses are designed to boost the realism of the generated image and the objects in the image respectively. However, using pooled feature as appearance representation means that both losses are location-insensitive, i.e., they only care about the presence/absence and strength of each learned semantic feature; much less about where the corresponding visual concept appear in the image. 

To address this problem, we propose to introduce an additional  appearance discriminator loss, which directly penalizes the spatial misalignment of each semantic feature between the generated and real images. Concretely, we use object feature maps' Gram matrix \cite{gatys2015neural} as a new appearance representation and feed it to the discriminator classification layer. Formally, we define the appearance of a generated object in the image as:
\vspace{-2mm}
\begin{equation}
    \mathcal{A}_{i} = \textbf{s}_{i}\textbf{s}_{i}^{T}/d_{s},
\end{equation}
where $d_{s}=C$ is the channel dimension of the feature map, $\textbf{s}_{i}\in \mathbb{R}^{C\times (H_{D} \times W_{D})}$ is the spatial dimension vectorized feature representation of the $i^{th}$ generated object in the image, computed as:
\begin{equation}
    \textbf{s}_{i} = \mathcal{R_{A}}(f_{im},b_{i}),
\end{equation}
where $\mathcal{R_{A}}(\cdot,\cdot)$ is the ROI align operator \cite{he2017mask}. For simplicity, the vectorization operation is omitted here.
The new appearance loss is then defined as:
\vspace{-2mm}
\begin{equation}
    \begin{aligned}
    \mathcal{L}_{a}(\mathcal{G},\mathcal{D}) = & \  \mathbb{E}_{\mathcal{A}^{r} \sim p_{data}^{r}(\mathcal{A}^{r})}[log(\mathcal{D}(\mathcal{A}^{r}|y)] \\
    & + \mathbb{E}_{\mathcal{A}^{g} \sim p_{data}^{g}(\mathcal{A}^{g})}[1-log(\mathcal{D}(\mathcal{A}^{g}|y)],
    \end{aligned}
\end{equation}
where $\mathcal{A}^{r}$ and $\mathcal{A}^{g}$ are the Gram matrices of object feature maps in real and generated images respectively, $y$ is their corresponding class label. More specifically, for the $i^{th}$ object in an image, its appearance loss is computed as:
\vspace{-2mm}
\begin{equation}
    \mathcal{D}(\mathcal{A}_{i}|y) =\frac{1}{C} \sum_{j=1}^{C}[\mathcal{A}_{i,j},\mathcal{E}(y_{i})]\mathcal{W}_{A},
\end{equation}
where $\mathcal{E}(y_{i})\in\mathbb{R}^{k}$ is the label embedding, and $\mathcal{W}_{A}\in \mathbb{R}^{(C+K)\times 1}$ is a linear layer. The Gram matrix here captures the correlation between different feature channels and is clearly location-sensitive: each entry only assumes a large value when the corresponding two features are both present and activated at the same location. This loss is thus complementary to the two conventional losses ($\mathcal{L}_{im}$ and $\mathcal{L}_{o}$) which emphasize the presence of the semantics only.  

\subsection{Training Objectives}

The final model is trained with the proposed appearance loss, together with image and object level losses \cite{zhao2019image,sun2019image}:
\begin{equation}
    \mathcal{G}^{*} = \text{arg} \ \underset{\mathcal{D}}{\text{min}} \ \underset{\mathcal{G}}{\text{max}} \ \ \mathcal{L}_{a}(\mathcal{G},\mathcal{D}) + \lambda_{im}\mathcal{L}_{im}(\mathcal{G},\mathcal{D}) + \lambda_{o}\mathcal{L}_{o}(\mathcal{G},\mathcal{D}),
\end{equation}
where $\lambda_{im}$ and $\lambda_{o}$ are the loss weight hyperparameters, and $\mathcal{L}_{im}$ and $\mathcal{L}_{o}$ are computed as:
\begin{equation}
    \begin{aligned}
    \mathcal{L}_{im}(\mathcal{G},\mathcal{D}) = & \  \mathbb{E}_{I_{im}^{r} \sim p_{data}^{r}(I_{im}^{r})}[\log(\mathcal{D}(I_{im}^{r})] \\
    & + \mathbb{E}_{I_{im}^{g} \sim p_{data}^{g}(I_{im}^{g})}[1-\log(\mathcal{D}(I_{im}^{g})], \\
    \mathcal{L}_{o}(\mathcal{G},\mathcal{D}) = & \  \mathbb{E}_{O^{r} \sim p_{data}^{r}(O^{r})}[\log(\mathcal{D}(O^{r}|y)] \\
    & + \mathbb{E}_{O^{g} \sim p_{data}^{g}(O^{g})}[1-\log(\mathcal{D}(O^{g}|y)],
    \end{aligned}
\end{equation}
where $I_{im}^{r}$ and $I_{im}^{g}$ are real and generated images respectively, and $O^{r}$ and $O^{g}$ are objects in the real and generated images.

\begin{table*}[t]
    \centering
    \small\addtolength{\tabcolsep}{0pt}
    \caption{Comparative results on COCO-Thing-Stuff and Visual Genome. E-D means encoder-decoder based generator, D means decoder-only based generator. $\dagger$ means improved decoder-only generator. 
    }
    \begin{tabular}{c|c|c| cc|cc|cc}
    \hline
    \multirow{2}{*}{Methods}& \multirow{2}{*}{Resolution} &  \multirow{2}{*}{Generator}&\multicolumn{2}{c|}{\textbf{Inception Score} $\uparrow$ } & \multicolumn{2}{c|}{\textbf{FID} $\downarrow$} &\multicolumn{2}{c}{\textbf{Diversity Score} $\uparrow$}\\
    \cline{4-9}
    & && COCO & VG &COCO &VG &COCO &VG\\
    \hline
    Real images  & 64 $\times$ 64 & - & 16.3 $\pm$ 0.4 &13.9 $\pm$ 0.5 &- & -& -&-  \\
    Real images  & 128 $\times$ 128 & - & 22.3 $\pm$ 0.5 &20.5 $\pm$ 1.5 &- &- &- &-  \\
    \hline
      \hline
    pix2pix \cite{isola2017image} & 64 $\times$ 64& E-D &3.5 $\pm$ 0.1 & 2.7 $\pm$ 0.02 & 121.97 & 142.86 & 0 & 0\\  
    Layout2im \cite{zhao2019image} & 64 $\times$ 64& E-D& 9.1 $\pm$ 0.1&8.1 $\pm$ 0.1 & 38.14&40.07 &0.15 $\pm$ 0.06 &0.17 $\pm$ 0.09  \\
    \hline
    Ours-ED & 64 $\times$ 64 &E-D& \textbf{10.27 $\pm$ 0.25} &\textbf{8.53 $\pm$ 0.13} & \textbf{31.32} & \textbf{33.91} & \textbf{0.39 $\pm$ 0.09} & \textbf{0.4$\pm$0.09 } \\
    \hline
    \hline
    Grid2Im \cite{ashual2019specifying} & 128 $\times$ 128 &E-D& 11.22 $\pm$ 0.15 & -& 63.44 & -& 0.28 $\pm$ 0.11 & - \\
    LostGAN-v1 \cite{sun2019image} & 128 $\times$ 128 &D &13.8 $\pm$ 0.4 &11.1 $\pm$ 0.6 &29.65 &29.36 &0.40 $\pm$ 0.09 &0.43 $\pm$ 0.09  \\
    LostGAN-v2 \cite{zhao2020layout2image} & 128 $\times$ 128 &$\text{D}^{\dagger}$&14.21 $\pm$ 0.4 &10.71 $\pm$ 0.76 &24.76 &29.00 &\textbf{0.55 $\pm$ 0.09} & 0.53 $\pm$ 0.09  \\
    OC-GAN \cite{sylvain2020object} & 128 $\times$ 128 &D& 14.0 $\pm$ 0.2 & 11.9 $\pm$ 0.5 & 36.04 & 28.91 &- &- \\
    AG-Layout2im \cite{ma2020attribute} & 128 $\times$ 128 &$\text{E-D}$& - & 8.5 $\pm$ 0.1 & - & 39.12 & - & 0.15 $\pm$ 0.09 \\
    \hline
    Ours-D & 128 $\times$ 128 &D& \textbf{15.62 $\pm$ 0.05} &\textbf{12.69 $\pm$ 0.45} & \textbf{22.32} &\textbf{21.78} & \textbf{0.55 $\pm$ 0.09} & \textbf{0.54 $\pm$ 0.09} \\
    
    \hline
    \end{tabular}
    \label{tab:auto_eval}
\end{table*}


\begin{figure*}[t]
\centering
\includegraphics[width=0.85\textwidth]{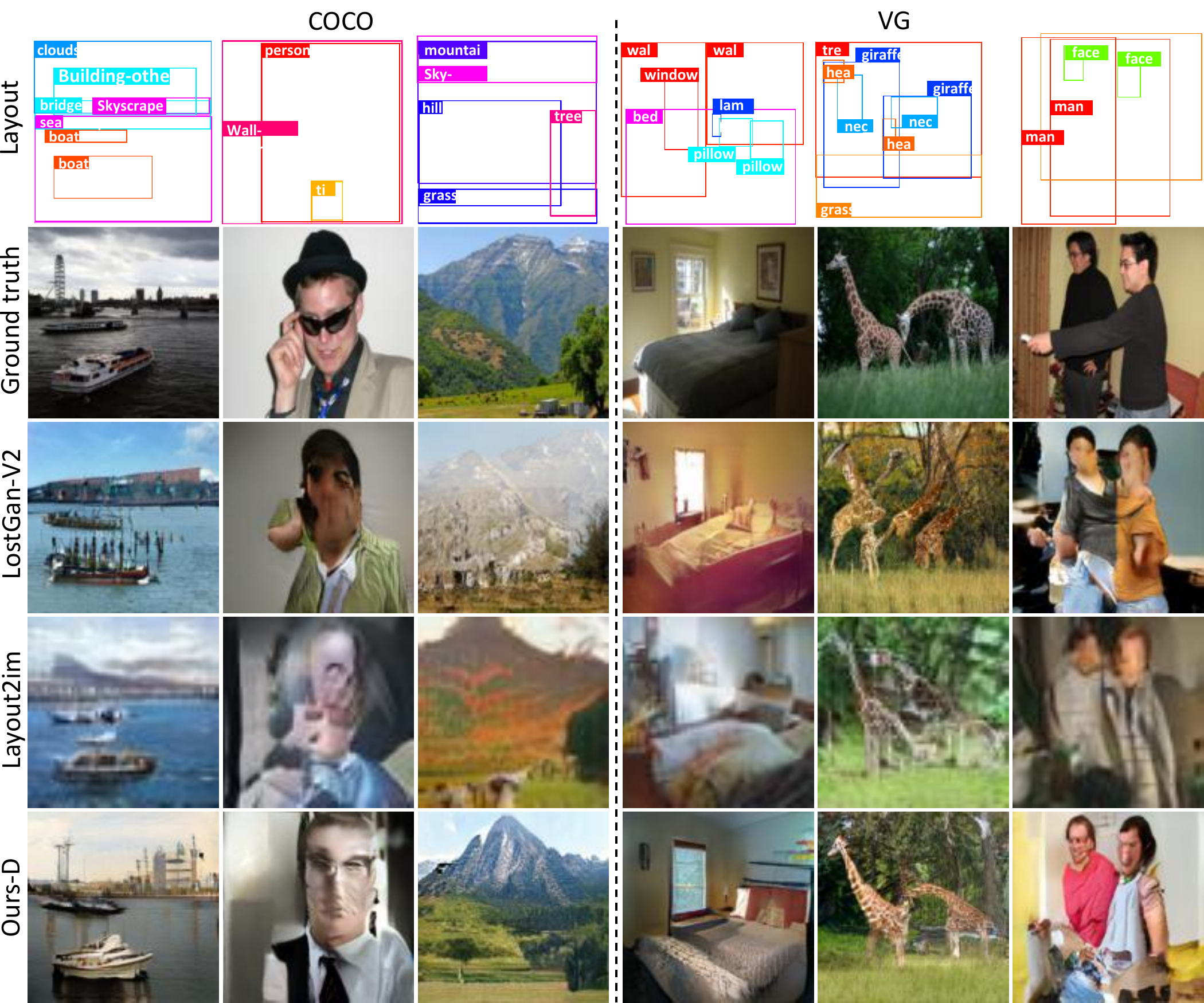}
\caption{Qualitative results comparing Ours-D against two  representative baselines  Layout2im \cite{zhao2020layout2image} and LostGAN-v2 \cite{sun2020learning}.}
\vspace{-10pt}
\label{fig:qualitative}
\end{figure*}

\begin{figure*}[t]
\centering
\includegraphics[width=0.9\textwidth]{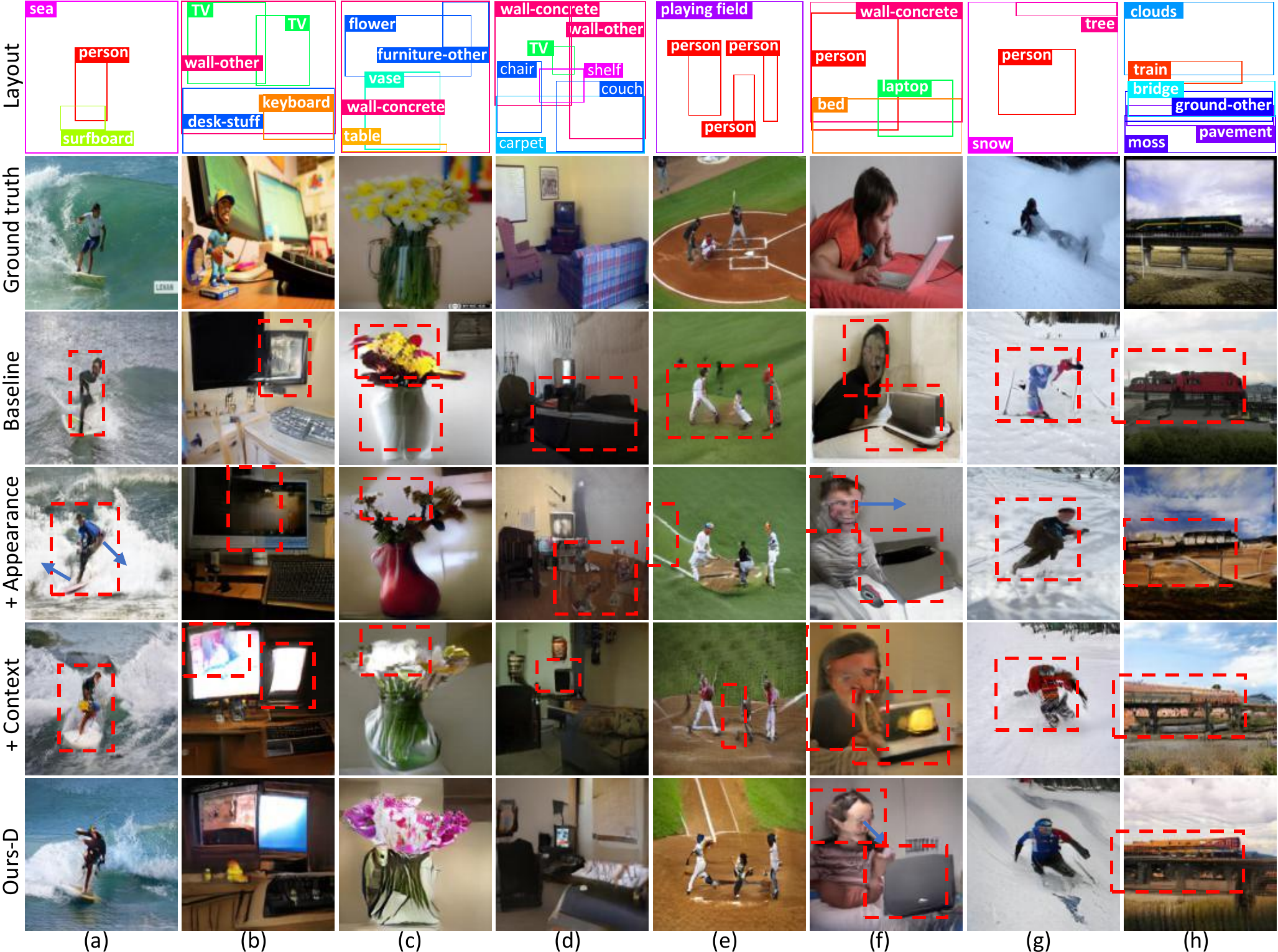}
\vspace{-0.2cm}
\caption{Qualitative ablative experimental results. Regions with clear generation quality differences are highlighted using red dashed boxes for close examination.}
\label{fig:ablative}
\end{figure*}

\begin{figure}[t]
\centering
\includegraphics[width=0.4\textwidth]{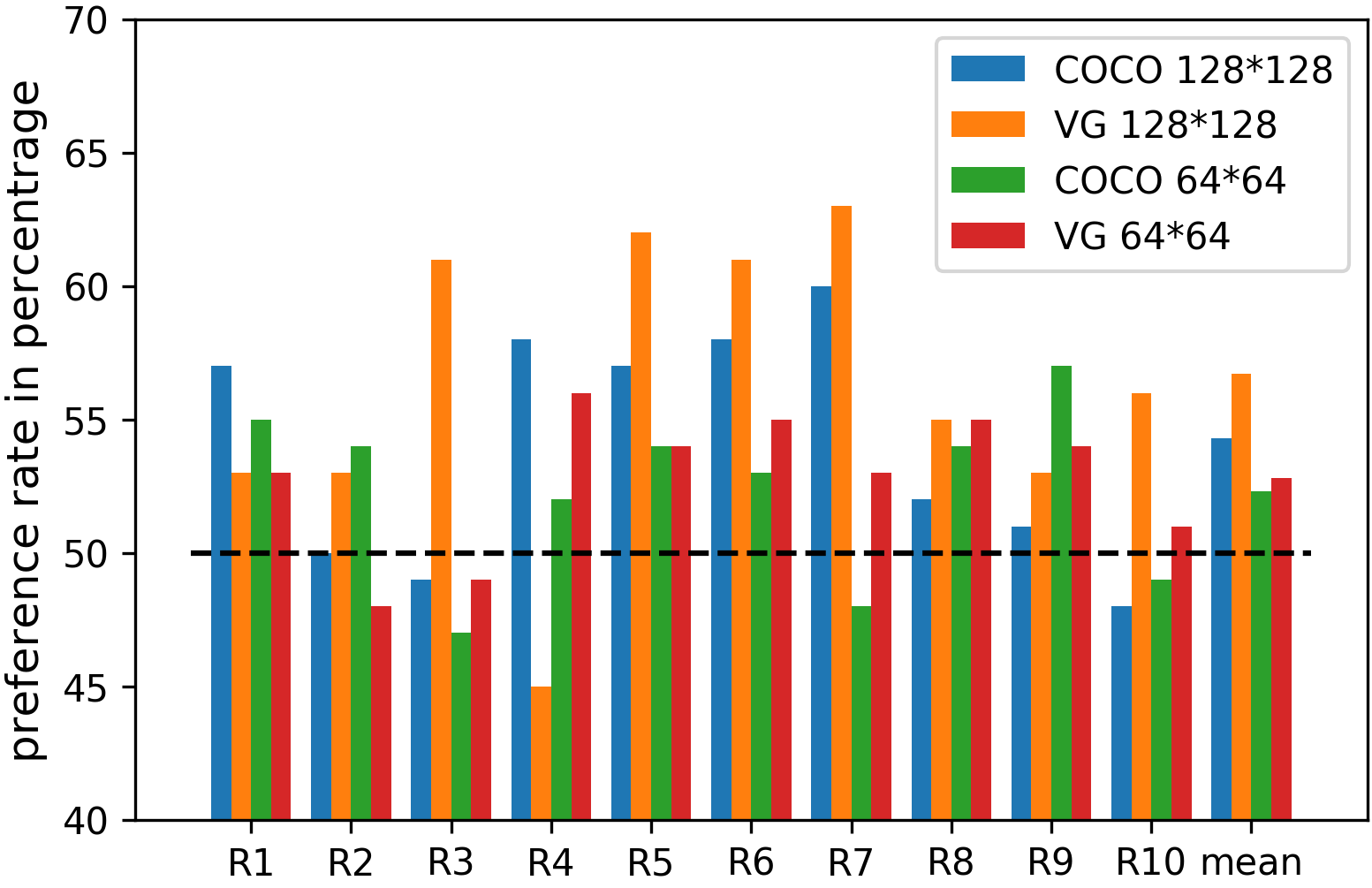}
\caption{The preference rate of our model.  A bar higher than dark dashed horizontal line indicates that our model is judged to be better than the compared baseline by the AMT workers.}
\vspace{-0.4cm}
\label{fig:amt}
\end{figure}

\section{Experiments}
\label{sec:exp}

\paragraph{Datasets} Two widely used benchmarks, COCO-Thing-Stuff \cite{lin2014microsoft,caesar2018coco} and Visual Genome \cite{krishna2017visual} are used in our experiments. COCO-Thing-Stuff includes bounding box annotations of the 91 \textit{stuff} classes in \cite{caesar2018coco} and the 80 \textit{thing/object} classes in \cite{lin2014microsoft}. Following \cite{zhao2019image,sun2019image}, only images with 3 to 8 bounding boxes are used in our experiments.
Visual Genome is originally built for complex scene understanding. The annotations in Visual Genome contain bounding boxes, object attributes, relationships, region descriptions, and
segmentation. As per standard in L2I generation, we only use the bounding boxes annotation in our experiments, and each layout contains 3 to 30 bounding boxes. We follow the splits in prior works \cite{zhao2019image,sun2019image} on both datasets to train and test our model.

\vspace{-0.4cm}
\paragraph{Implementation Details}
Our model is implemented with PyTorch. 
To show the general applicability of our proposed method, and for fair comparison with prior works, we adopt both encoder-decoder and decoder-only generators in the two instantiations of our method (termed Ours-ED and Ours-D respectively). The encoder-decoder generator has the same architecture as used in \cite{zhao2019image}, and the decoder-only generator shares the same architecture as used in \cite{sun2019image}. Following \cite{zhao2019image,sun2019image}, the resolution of generated images is $64\times64$ for the encoder-decoder generator and $128\times128$ for the decoder-only generator. The learning rate is set to $1e^{-4}$ for both generator and discriminator in all the experiments. 
We train our model for 200 epochs. The loss weight hyperparameters $\lambda_{im}$ and $\lambda_{o}$ are set to 0.1 and 1, respectively.

\vspace{-0.4cm}
\paragraph{Evaluation Metrics} 
We evaluate our model both automatically and manually. In automatic evaluation, we adopt three widely used metrics, namely Inception Score \cite{salimans2016improved}, Fr\'{e}chet Inception Distance (FID) \cite{heusel2017gans} and Diversity Score \cite{zhang2018unreasonable}. Inception Score evaluates the quality of the generated images. FID computes the statistical distance between the generated images and the real images. Diversity Score compares the difference between the generated image and the real image from the same layout. Following prior evaluation protocol \cite{ashual2019specifying}, for each layout, we generate five images in COCO-Thing-Stuff and one image in Visual Genome.  In manual evaluation, we run perceptual studies on Amazon Mechanical Turk (AMT) to compare the quality of the generated images from different models. Ten participants engaged in the evaluation. Each participant was given 100 randomly sampled layouts from the testing dataset as well as the corresponding generated images from  different models. All participants were asked to vote for their preferred image according to the image's quality and the matching degree to the paired layout. 
We compute the preference rate of each model from all participants. Due to the difference in generated image's resolution and for fair comparison, we compare our encoder-decoder generator based instantiation (Ours-ED) with the state-of-the-art encoder-decoder generator based baseline Layout2im \cite{zhao2019image} and decoder-only instantiation (Ours-D) with the state-of-the-art decoder-only generator based baseline LostGAN-v2 \cite{sun2020learning}. 
In both comparisons, the generated images are of the same resolution. 

\vspace{-0.4cm}
\paragraph{Main Results}
We compare our method with existing L2I models \cite{zhao2019image,sun2019image,sun2020learning,sylvain2020object,ma2020attribute}, the pix2pix model \cite{isola2017image} which takes the input feature maps constructed from layout as implemented in \cite{zhao2019image}, and the Grid2Im model \cite{ashual2019specifying} which receives scene graph as input. The following observations can be made on the quantitative results shown in Table~\ref{tab:auto_eval}. 
\textbf{(1)} Our method outperforms all compared methods on all benchmarks with both architectures and under all three automatic evaluations metrics, particularly for Inception Score and FID. \textbf{(2)} The more recent L2I methods take a decoder-only generator. Taking the same architecture but with the two new components, our method (Ours-D) achieves new state-of-the-art. Fig~\ref{fig:amt} shows detailed statistics in the human evaluation on AMT. Among all 40 evaluation sets, our model won 32 sets. The preference rate is clearly higher at the higher resolution with more complex images (i.e., $128 \times 128$, VG dataset). 
Some qualitative results are shown in Fig.~\ref{fig:qualitative}. It is evident from these examples that the images generated using our method are much more context-aware, i.e., different objects co-exist in harmony with each other and the background. Importantly, each generated object has sharper texture, clearer shape boundary with respect to background inside the object bounding box, and overall much more spatially-coherent than those generated by existing L2I models. 
\vspace{-0.4cm}

\paragraph{Ablation Study}\label{ablation}
In this experiment, we adopt LostGAN-v1 \cite{sun2019image} as our baseline and evaluate the effects of introducing  our context transformation module and location-sensitive appearance representation. The quantitative results are shown in Table~\ref{tab:ablative}. We can see that both our context-aware feature transformation module and new appearance representation improve the baseline significantly on their own and when combined give a further boost. Some qualitative results are shown in Fig.~\ref{fig:ablative}. It is clear that the model trained with our appearance representation can generate objects with much better appearance both in terms of shape and texture (TV in Fig.~\ref{fig:ablative}(b) and person in Fig.~\ref{fig:ablative}(a)(f)(g)). Context transformation also plays an important role: the generated occluded regions become more natural (Fig.~\ref{fig:ablative}(b)(f)); each object's pose is also more in-tune with its surrounding  objects and background, e.g.~the surfing person's body pose is more physically plausible in Fig.~\ref{fig:ablative}(a); so is the person's head pose in the presence of the laptop in Fig.~\ref{fig:ablative}(f).  
\vspace{-0.4cm}

\begin{table}[t]
    \centering
    \caption{Ablation study on COCO-Thing-Stuff dataset.}
    \small\addtolength{\tabcolsep}{-2pt}
    \begin{tabular}{ccc|cc}
    \hline
    baseline \cite{sun2019image} & context & appearance & \textbf{Inception Score}& \textbf{FID} \\
    \hline
    \checkmark& & & 13.8 $\pm$ 0.4 & 29.65\\
    \checkmark & \checkmark & & 14.97 $\pm$ 0.27 &24.05\\
    \checkmark &  & \checkmark &  15.28 $\pm$ 0.24 & \textbf{21.73}\\
    \checkmark & \checkmark & \checkmark & \textbf{15.62 $\pm$ 0.05} & 22.32\\
    \hline
    \end{tabular}
    \label{tab:ablative}
    \vspace{-10pt}
\end{table}


\begin{figure}[h]
\centering
\includegraphics[width=0.45\textwidth]{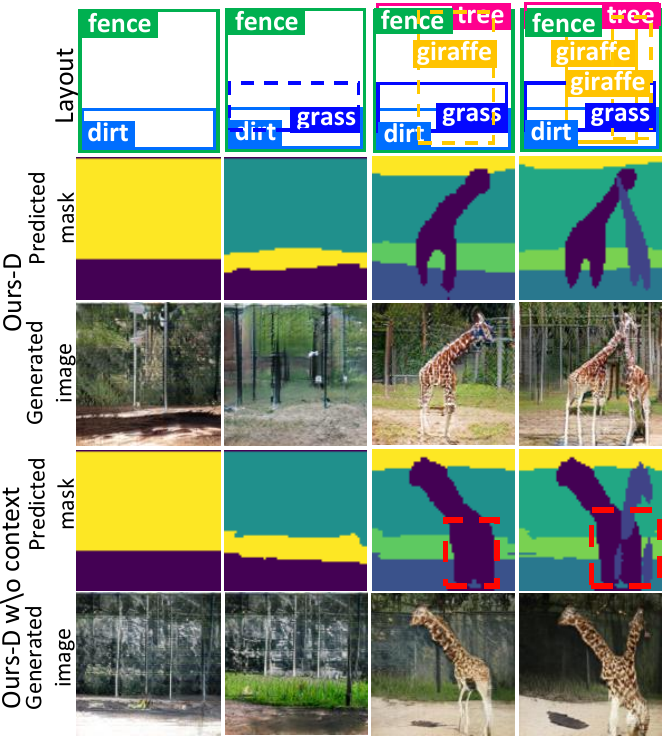}
~
\caption{\small Qualitative examples about the contribution of context transformation in the complex scene generation. From left to right, at each time, we add one more bounding box into the previous layout, visualizing the predicted masks as well as the generated image by a model with our context transformation (Ours-D), and the same model without context transformation. Regions to pay more attention to are highlighted in dashed boxes.}
\vspace{-0.4cm}
\label{fig:ablative_mask}
\end{figure}


\paragraph{How Our Context Transformation Module Works}
In the decoder-only generator, a mask is generated using the representation of each bounding box to predict the fine-grained shape or structure of the object in each bounding box (Eq.~\ref{eq:mask}). Without the context information in the feature representation, the generated masks would interfere with each other. This could result in irregular or incomplete object shape particularly in the occluded regions, which would further affect the feature modulation defined in Eq.~\ref{eq:c_mask}. We investigate this effect by adding more bounding boxes into a layout, and visualizing the predicted masks as well as the generated images. The visualization results in Fig.~\ref{fig:ablative_mask} show clearly that the context-aware feature transformation module reduced the negative inter-object appearance interference in a complex scene when occlusion exists, yielding better appearance for the generated objects.


\section{Conclusion}

In this work, we proposed a novel context feature transformation module and a location-sensitive appearance representation to improve existing layout to image (L2I) generation models. In particular, they are designed to address existing models' limitations on lacking context-aware modeling in their generator and spatially sensitive appearance representation in their discriminator. Extensive experiments demonstrate the effectiveness of our method, yielding new state-of-the-art on two benchmarks.

\section*{Acknowledgment}
This work was supported by the Federal Ministry of Education and Research (BMBF), Germany under the project LeibnizKILabor(grant no.01DD20003) and the Deutsche Forschungsgemeinschaft  (DFG)  under  Germany’s  Excellence  Strategy  within  the  Cluster of Excellence PhoenixD (EXC 2122).

{\small
\bibliographystyle{ieee_fullname}
\bibliography{egbib}
}

\end{document}